%% file: main.tex
\documentclass{article}

\usepackage[preprint]{neurips_2024}

\usepackage[utf8]{inputenc} 
\usepackage[T1]{fontenc}    
\usepackage{hyperref}       
\usepackage{url}            
\usepackage{booktabs}       
\usepackage{amsfonts}       
\usepackage{nicefrac}       
\usepackage{microtype}      
\usepackage{soul}           
\usepackage[usenames,dvipsnames]{xcolor}         
\usepackage{import}         
\usepackage{multirow}
\usepackage{amsmath}
\usepackage{breqn}
\usepackage{natbib}
\usepackage{csquotes}
\usepackage{alltt}

\usepackage{marginnote}

\usepackage{graphicx}

\DeclareMathOperator*{\argmax}{arg\,max}

\author{%
  Eric Yeh, John Cadigan, Ran Chen, Dick Crouch, Melinda Gervasio, Dayne Freitag \\
  SRI\\
  Menlo Park, CA 94025, USA \\
  \texttt{eric.yeh|john.cadigan|ran.chen|dick.rouch|melinda.gervasion|dayne.freitag@sri.com} \\
}

\begin{document}

\title{Interpolative Decoding: Exploring the Spectrum of Personality Traits in LLMs}
\maketitle
\begin{abstract}
Recent research has explored using very large language models (LLMs) as proxies for humans in tasks such as simulation, surveys, and studies.
While LLMs do not possess a human psychology, they often can emulate human behaviors with sufficiently high fidelity to drive simulations to test human behavioral hypotheses, exhibiting more nuance and range than the rule-based agents often employed in behavioral economics.  One key area of interest is the effect of personality on decision making, but the requirement that a prompt must be created for every tested personality profile introduces experimental overhead and degrades replicability.  To address this issue, we leverage \emph{interpolative decoding}, representing each dimension of personality as a pair of opposed prompts and employing an interpolation parameter to simulate behavior along the dimension.  We show that interpolative decoding reliably modulates scores along each of the Big Five dimensions.  We then show how interpolative decoding causes LLMs to mimic human decision-making behavior in economic games, replicating results from human psychological research.  Finally, we present preliminary results of our efforts to ``twin'' individual human players in a collaborative game through systematic search for points in interpolation space that cause the system to replicate actions taken by the human subject.  
\end{abstract}

\import{sections}{introduction}

\import{sections}{background}

\import{sections}{idecoding}

\import{sections}{twinning}

\import{sections}{experiments}

\import{sections}{discussion}

\section{Conclusion}

With their ability to emulate human subjects, generative LLMs offer intriguing options for psychology and social science research.  Under the state of the art, such emulation requires either tuning LLMs on significant amounts of data or optimizing prompts to match individuals or personality types---requirements that reduce the usability of these techniques and the replicability of reported outcomes.  Interpolative decoding addresses these problems, requiring only a handful of prompts to define a parametric behavioral space, and offering a mechanism by which any point in the space can be emulated.  We have shown, using standard models of personality, that interpolative decoding gives rise to decision making behaviors consistent with intuition and previous research.  And we have provided preliminary evidence, in a process called \emph{twinning}, that the decision making of individual subjects can be emulated.  While we do not know with what fidelity twinning can ultimately be performed, we submit that a research agenda centered on maximizing fidelity offers new possibilities for interrogating existing theories of decision making.

\section{Acknowledgments}
This material is based upon work supported by the Defense Advanced Research Projects
Agency (DARPA) under Contract No. HR00112490411. Any opinions, findings and
conclusions or recommendations expressed in this material are those of the author(s)
and do not necessarily reflect the views of the DARPA.  Distribution Statement A. Approved for public release; distribution is unlimited.

\bibliography{main}
\bibliographystyle{plainnat}

\import{appendices}{appendix1}

\end{document}

%% file: sections/introduction.tex
\section{Introduction}
\label{sec:introduction}

As generative large language models (LLMs) have grown in size, their language has become increasingly human, motivating an interest in emulating human \emph{behaviors} for a variety of applications.
As LLMs are trained on web-scale examples of human behavior, one could argue they also exhibit similar behaviors.  Indeed, recent research offers demonstrations that LLMs reproduce human biases, both social~\cite{frisch_llm_2024,jiang_personallm_2024} and cognitive~\cite{jones2022capturing}; that they exhibit plausible emergent behaviors in simulated social settings~\cite{park2023generativeagentsinteractivesimulacra}; and that, applied at scale, they can reproduce the responses of human populations to surveys or polls~\cite{manning_automated_2024}.  An arguably important use of LLMs is to illuminate the factors that bear on human \emph{decision making}, supplementing the rule-based models that fields such as behavioral economics rely on~\cite{moffittbehavioral}.


Previous psychology research has identified some of these decision factors. 
For example, traits from personality models such as Big Five \cite{McCrae1992} and HEXACO \cite{bigfive, hexaco} have been shown to impact decision making in economic games \cite{g9020030, thielmann2020personality}, and ultimately team performance in collaborative settings \cite{hexaco_teamqork_zahl_2025}.
It is also known that capacity to integrate information from multiple sources, and the manner in which that information is integrated, impacts decision making.
Humans have demonstrated variability in weighing information from different sources, such as from social cues versus personal experiences \cite{parnamets2020integration}.  
Weighting of these factors may be due to psychological factors \cite{perceptualcueweighting_yu_2022} or this may even be physiological \cite{ou2023individual}.

One impediment to the use of LLMs to study these questions is the need to create a specific ``prompt'' (the conditioning input to the LLM) to emulate any particular personality profile.  
Empirically motivated models of human psychology, such as Big Five and HEXACO, posit spectra along which personality traits vary.  To emulate a personality type at a particular point in a spectrum using a generative LLM, one must prompt the LLM with a full description of that point.  This requirement represents a mere nuisance if the objective is only to emulate a chosen multi-factor personality profile, but if the objective is to \emph{find} the personality profile that best matches a particular individual's decision making, it is a major obstacle.

In this work, we describe how \emph{interpolative prompt decoding} can be used to represent intermediate points along character spectra, specifically those concerned with personality and information integration.\footnote{We use \emph{character} as a superordinate category encompassing personality, cognitive style, and any other identifiable factor on which behavior might be conditioned.}  This type of algorithm combines and modulates the contribution of multiple output distributions and has been used to influence LLM outputs, from surfacing biases in gendered language \cite{yona2023surfacingbiaseslargelanguage} to activating different personality traits \cite{li-etal-2025-big5}.  We hypothesize that interpolation between trait extrema prompts should result in LLM outputs that mimic human ones along that trait spectrum.  In particular, we contend LLMs responses can mimic those of humans' on psychological inventories and in decision-making tasks governed by those traits.
Similarly, we expect similar behavior when varying different sources of information to the LLM.

We note that ``behavior'' means first and foremost communicative behavior, inasmuch as these are language models, but we are ultimately interested in decision making and task performance.  To that end, we adopt the standard trick of requiring the models to render decisions about task actions in a semi-structured format suitable for easy downstream processing and scoring.



Having a parameterized model of how personality governs behavior implies the inverse, can these parameters be inferred from behavior?
Or more precisely, if we can affect LLM behavior by modulating the contribution of different psychological dimensions,  we should be able to \emph{twin} a human through observation by tuning relevant traits and characteristics so the LLM behavior matches the human's.

Given these hypotheses we pose the following research questions (RQ):

\textbf{RQ 1: Psychological Soundness}. \emph{Does interpolative decoding over generally trained LLMs produce behaviors consistent with standard models of personality?}

\textbf{RQ 2: Decision Making}. \emph{What impact does personality modulation, using interpolative decoding, have on decision making?}

\textbf{RQ 3: Human Twinning}. \emph{Using interpolative decoding, how well can we emulate the decision making of individual human subjects?}

To examine these questions, we present several experiments.  The first examines LLM responses as scored by the Big Five personality inventory.
The next set of experiments examines how interpolative prompt decoding impacts decision-making using the dictator game, an economic game shown to correlate well with foundational prosocial behaviors for effective collaboration and teamwork.
We then examine how interpolative decoding changes decisions and reasoning made by an LLM when reweighting different sources of information.  
To address twinning, we present the \emph{Facsimile of Intelligent Life} (FOIL) system and our initial efforts to ``twin'' human players of a collaborative game, by systematically searching through multidimensional interpolation space. 
We follow up with a discussion and highlight future directions.
Source code and data for our experiments will be made available at \url{https://github.com/SRI-AIC/foil}.


%% file: sections/background.tex
\section{Background}
\label{sec:background}

Big Five \citep{bigfive} and HEXACO \citep{hexaco} are personality factor models predicated on the hypothesis that important aspects of personality are reflected in language use and derived from studies of Asian and European languages.
Big Five posits that human personality is composed of five orthogonal factors: \emph{openness}, \emph{conscientiousness}, \emph{extraversion}, \emph{agreeableness}, and \emph{neuroticism} \citep{Goldberg1990, bigfive}.  HEXACO extends Big Five with the \emph{honesty-humility} factor, which assesses sincerity, fairness, modesty, and greed.
Previous research has investigated the connection between HEXACO traits and the prosocial behavior deemed foundational for effective teams.
For example, HEXACO traits associated with prosocial behavior, such as honesty-humility and agreeableness, have been shown to correlate with the success and efficiency of software development teams \citep{hexaco_teamqork_zahl_2025}. 

Based on their responses to personality assessment inventories, different LLMs exhibit distinct ``innate'' profiles, with GPT 4 and Llama2, for example, scoring relatively highly on agreeableness and neuroticism, respectively~\citep{sorokovikova_llms_2024}, but can successfully emulate arbitrary personality profiles when prompted with simple characterizations of the desired profile~\citep{frisch_llm_2024,jiang_personallm_2024}. Most similar to our work is BIG5-CHAT \citep{li-etal-2025-big5}, which aimed to modify LLM output to match a desired Big-Five personality profile.  However, that work focused on the extremes of Big-Five personality traits and required LLMs fine-tuned to represent those extremes.  Our approach requires no fine tuning and involves only a single prompt-driven model, making it suitable for use with broadly available LLMs, such as those in the GPT family.   

The demonstrated ability of LLMs to ``channel'' human social actors lies behind a growing interest in using LLMs as human surrogates in social science research, much of it focused on broad types and population-level effects. 
\citet{grossmann_ai_2023} considers some of the ways in which ``agent-based modeling'' might facilitating social science research, arguing that it can be used to accelerate data collection and assist with early-stage experiment design.
\citet{kim_ai-augmented_2024} exemplifies AI-assisted experiment design, reporting good accuracy in ``unasked opinion prediction," predicting responses to questions omitted from public attitude surveys.
\citet{ashokkumar_predicting_2024} reports substantial success in reproducing the results of social science surveys \emph{in silico}, using GPT-4 to emulate individual respondents based on prompts that describe demographic profiles.  
\citet{manning_automated_2024} describes a framework for LLM-driven design, implementation, and execution of small-scale economic experiments involving simulated participants.

Our work on interpolated decoding is based on previous research seeking effective ways to influence LLM outputs that do not involve training the LLM. For example, \emph{contrastive decoding} was originally proposed as a means to improve the quality of generated language \citep{liu-etal-2021-dexperts, li2023contrastivedecodingopenendedtext}, and then used as a means to expose social biases latent in LLMs \citep{yona2023surfacingbiaseslargelanguage}.  Variants of these techniques have since found other uses, such as ensuring that LLMs reflect the preferences of their individual users~\citep{bo_steerable_2025} and tailoring the language they generate to particular types of human consumers~\citep{he_context_2025}.

Economic games that pose scenarios presenting a handful of distinct choices are commonly used to investigate the link between personality and decision making.  
For example, a player of the dictator game, an instrument for gauging prosocial behavior, is asked to divide \$100 between themselves and a coworker.
Recent studies have observed positive correlations high dictator game payouts and HEXACO traits honesty-humililty and (to a lesser extent) agreeableness \citep{g9020030, thielmann2020personality}.

%% file: sections/idecoding.tex
\section{Interpolative Decoding}
\label{sec:idecoding}

Generative LLMs are adept at playacting, a direct result of the procedure used to train them, which we sketch here.  After a pretraining phase, in which a model acquires language fluency through exposure to large volumes of text, they are further trained to perform useful language generation tasks by fitting to a corpus of prompts paired with desired responses.  Prompts often involve a preamble that describes the type of character the model is expected to emulate.  Much of the utility and versatility of these models is a consequence of their ability to ``channel'' stipulated language user types based on a description, an ability that personification research seeks to increase.

The fidelity of generated text to these character specifications is an important question, one often answered anecdotally.  However, rigor is possible in domains, such as personality profiling, that offer textual assessment instruments.  Of course, while it is straightforward to author and optimize prompts that describe the extremes of individual personality factors, it is less clear how to approximate intermediate settings---the settings most human subjects occupy.

\begin{figure}
\centering
\includegraphics[width=1.0\linewidth]{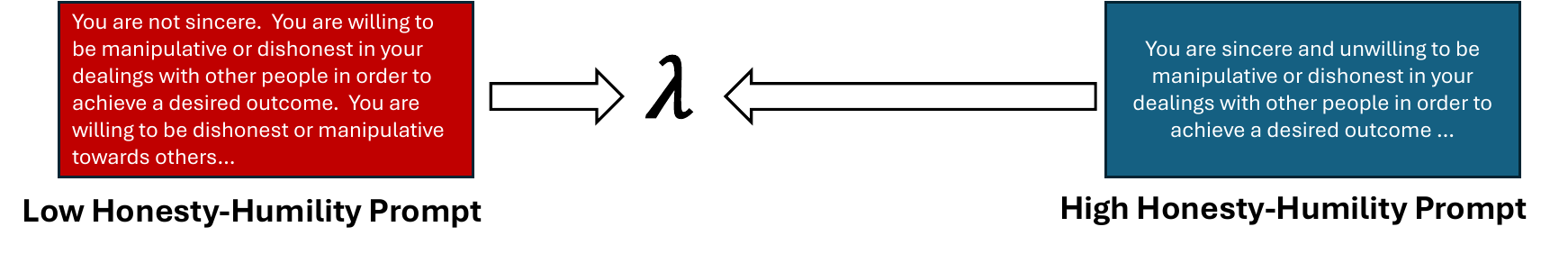}
\caption{\label{fig:conops} Interpolative decoding enables a generative model to approximate behavior between two character extremes based on fixed textual descriptions.}
\end{figure}

Interpolative decoding is intended to address this need.  As shown in Figure~\ref{fig:conops}, a spectrum of character (here the honesty-humility factor from HEXACO) is represented by a pair of textual descriptions, each describing an extreme along the given spectrum.  The technique relies on a real-valued interpolation parameter $\lambda$, the value of which controls where on the spectrum the desired character lies.  The hope is that points along the spectrum can be approximated more conveniently and comprehensively than through the creation of point-specific prompts.

Interpolative decoding is performed in the context of auto-regressive text generation. The core idea is to obtain next-token output probabilities from LLMs conditioned on the two endpoint prompts and then use an averaging strategy governed by $\lambda$ to obtain the final next-token distribution. 

In this work, we explore two mechanisms for interpolating: \emph{mixture} and \emph{contrastive}.  Mixture decoding creates a next-token distribution ($P'$) using a weighted average of the output distributions from the model conditioned solely on Prompt A ($P_A$) and solely on Prompt B ($P_A$):

\begin{equation}
    P'(t)=\frac{1}{Z}(\lambda P_A (t) + (1 - \lambda) P_B (t))
\end{equation}

With Z acting as the partition to ensure $P'$ is a distribution.

Contrastive decoding is an alternative that amplifies the distributional differences between $P_A$ and $P_B$:

\begin{equation}
    P'(t)=\textrm{softmax}(P_A(t) + \lambda (P_A(t)-P_B(t))
\end{equation}

Note that the contrastive formulation is asymmetric, ``anchoring'' off one of the prompt-conditioned next-token distributions ($P_A (t)$).  However, in practice we have found no reason to prefer as anchor one spectrum end point over the other; we are generally able to produce effective character interpolation with either choice.

%% file: sections/twinning.tex
\section{Twinning}
\label{sec:twinning}

We are interested not just in demonstrating that interpolative decoding can be used to replicate results from research into personality---results typically expressed as averages over cohorts---but also in using these techniques to emulate the communicative or decision making behavior of individuals.  In a process we call \emph{twinning}, we observe a subject's behavior in a constrained domain amenable to LLM emulation and seek to configure an AI system in a way that maximizes \emph{fidelity} to the subject, i.e., the tendency of the system to make the same decisions as the subject.

The potential uses of twinning are many.  If it can be accomplished with high fidelity, it offers a basis for conducting \emph{in silico} experiments that are either unethical or infeasible to perform with human subjects, e.g., experiments involving more iterations than a human subject can sustain.  But imperfect fidelity, too, provides opportunities for new insights.  For example, we might stipulate a dimension of character omitted by Big Five, one putatively important for predicting behavior.  If the dimension can be characterized by extremal prompts, we can assess its inclusion in the character model.  Increases in fidelity serve as evidence that the posited dimension complements Big Five---at least in the behavioral domain under study.

We can frame twinning as an optimization task.  Given the low and high trait descriptors $\tau_l, \tau_h$, a framing scenario $s$, and a sample of observed behavior $B = \{ (o_i, a_i) \}$ presented as a set of observation-action pairs, we search for a value of the interpolation parameter $\lambda$ most likely to cause the LLM $L$ to exhibit $B$:

\begin{equation}
    \argmax_{\lambda} \sum_{(o_i, a_i) \in B} P(a_i | o_i, \lambda, \tau_l, \tau_h, s, L)
    \label{eqn:lambda_argmax}
\end{equation}


If we have reason to suppose that the sum in Equation~\ref{eqn:lambda_argmax} is unimodal in $\lambda$, we can find a maximizing $\lambda$ efficiently with the Golden Section method~\cite{kiefer1953sequential}.  However, while the assumption of unimodality may be valid for simple economic games, there is no reason to suppose that it holds for arbitrary action spaces and, especially, optimization across multiple dimensions (e.g., all Big Five dimensions simultaneously).  Furthermore, evaluating the sum in Equation~\ref{eqn:lambda_argmax} is comparatively expensive, inasmuch as it involves multiple invocations of interpolative decoding for each setting of $\lambda$.  We can mitigate this expense somewhat with prompt caching\footnote{https://platform.openai.com/docs/guides/prompt-caching}, a technique that saves LLM encoding expense by processing invariant parts of an iteratively varied prompt only once, but this approach does nothing to limit the number of $\lambda$ values we must consider.


One approach to reduce the computational expense of sampling is to train a regressor to identify the value of $\lambda$, given the extrema prompts, the LLM, and the response it provided.
This mechanism can also provide a sufficiently close starting candidate $\lambda$, to reduce the number of search points to evaluate when running a search procedure.
In order to reduce the number of points to evaluate when twinning, we pre-train a regressor to produce a sufficiently close candidate $\lambda$ to begin the search from.  
Given a LLM to twin from, $L$, we develop a regression dataset by running interpolative decoding against a variety of scenarios $s \in S$ and candidate observations $o \in O$ against the desired trait extrema $\tau_l, \tau_h$ at several values of $\lambda$.
More formally, we create the regression dataset by recording the action $a$ produced by running interpolative decoding on the LLM, $\mathrm{ID}(o,\lambda,\tau_l,\tau_h,s,L)\rightarrow a$, where the function $\mathrm{ID}$ assembles the necessary low and high input prompts from $s, o$ and $\tau_l, \tau_h$.

In this work we explore the ability of sampling $\lambda$s and traits to produce a probability distribution of actions that best explain the actions performed by a human playing a task.  We also present an initial exploration of regressing $\lambda$ using LLM generated data.

An immediate future direction is to borrow ideas shown to work for the problem of hyperparameter optimization~\cite{hyper_opt}, a concern closely isomorphic to ours, in which the objective is to find values for training parameters (\emph{hyperparameters}) that maximize performance of an opaque machine learning algorithm, the training of which entails considerable expense.  In this context, Bayesian optimization under Gaussian priors is a principled approach with demonstrated effectiveness~\cite{frazier2018tutorial}.






%% file: sections/experiments.tex
\section{Experiments}
\label{sec:experiments}

In this section we describe experiments designed to answer the research questions articulated in Section~\ref{sec:introduction}.  

\import{experiments}{psycho_soundness}

\import{experiments}{decision_making}



\import{experiments}{exp_twinning}


%% file: sections/experiments/psycho_soundness.tex
\subsection{Psychological Soundness}
\label{sec:psycho_soudness}

We present experiments which evaluate the degree to which an LLM can be prompted to answer at a specific level for each of the Big Five traits based on a \textit{descriptive prompt prefix} and \textit{few-shot examples}. To make the experiment more robust, we partitioned few-shot examples and assessment examples according to Big Five facets which were identified by the inventory creators; for example, the 6 facets of conscientiousness are competence, order, dutifulness, achievement striving, self-discipline, and deliberation. We pick 3 facets for examples and 3 for assessment, resulting in an even-split of 12 and 12. We randomly answer the few shot examples which also creates the target for the test. To create our descriptive prompt prefix, we split the possible distribution of answers into thirds and describe the LLM as low, middle or high in a particular trait. We use the description appearing in the results page of the test creators website to populate the prompt and describe the trait. 

With our prompt constructed, we then collected LLM responses to the remaining facets for the trait. Just as a human test taker, the LLM is solicited responses to inventory questions, ranging from strongly disagree to five strongly agree. These questions are given a Likert scale, which is ascending or descending based on the question.  For example, a high conscientiousness personality would answer 'Strongly agree' to the question, ``\emph{I am someone who likes to tidy up.}'' (facet 2), resulting in a score of 5 being added to that trait. Other questions such as ``\emph{Jump into things without thinking}`` (facet 6) are descending for the same trait. From these responses, we average scores to get one input and output trait score. Repeating this experiment across facets, we gather 20 such pairs of scores based on all combinations of facets ($\binom{6}{3}$) for a given  and can calculate how well input and output examples correlate.


Here, we configure interpolative decoding to align with individual dimensions posited by models of personality and investigate whether by varying $\lambda$ we observe changes in behavior consistent with the personality model. Here we used the same sort of prompt from before, but contrasted it with a "neutral prompt" in which the few-shot examples were not well correlated with specific personality traits. If interpolative decoding is well behaved, we should see that the LLM's answers to questions change in a way consistent with the model.  When tested on extraversion at a high $\lambda$ value, it should receive a high extraversion score.

\begin{table}[t]
    \centering
    \begin{tabular}{|c|c|c|c|c|}
        \hline
         \textbf{Openness} &  \textbf{Conscientiousness} & \textbf{Extraversion} & \textbf{Agreeableness} & \textbf{Neuroticism} \\
         \hline
         0.69 & 0.83 & 0.83 & 0.70 & 0.73 \\
         \hline
    \end{tabular}
    \caption{Spearman correlations between LLM answers for Big Five inventory and targeted Big Five levels.}
    \label{tab:bigfive_correlations}
\end{table}

\begin{figure}[ht]
    \centering
    \includegraphics[width=1.0\linewidth]{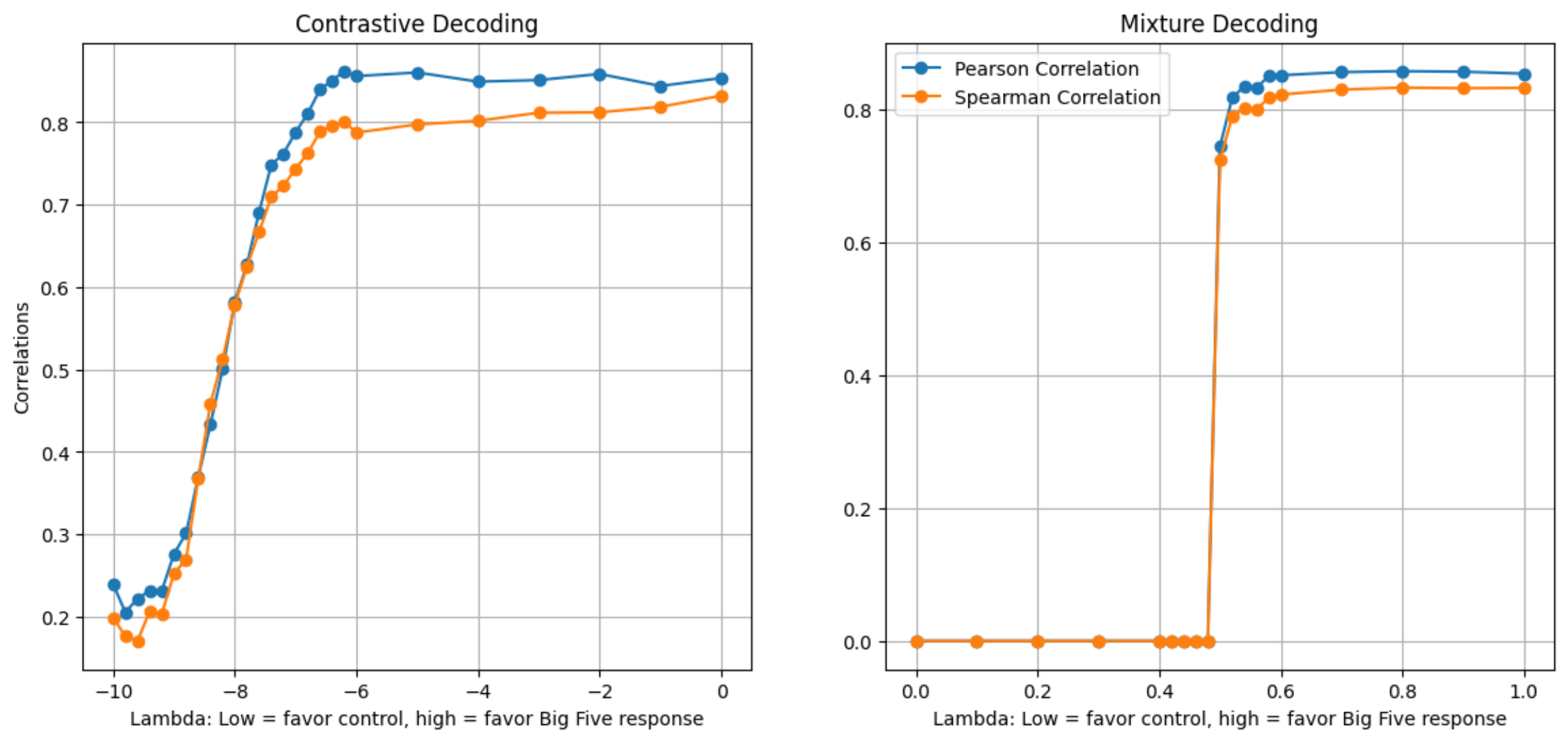}
    \caption{Correlations with Big Five scores between control (low lambda) and Big Five (high lambda) prompts, for contrastive (left) and mixture decoding (right).  }
    \label{fig:bigfive_corrs}
\end{figure}

We systematically varied $\lambda$ across a range of values across which we had previously observed shifts in language and behavior.  The two decoding methods, contrastive and mixture, have different ranges. The effective range for contrastive decoding is $-10$ to $0$, while mixture decoding used $0$ to $1$. For the purpose of interpolative decoding, each of these trait prompts was paired with a control prompt having no relation to the target trait.  For example, a control prompt might include the statement, ``\emph{You believe that pineapple belongs on pizza}.''
The full set of control prompts are presented in Appendix \ref{appendix:control_prompts}.

From these responses, we scored for each trait and computed the Pearson correlation between $\lambda$ and trait score, as shown in Table~\ref{tab:bigfive_correlations}.  
Figure~\ref{fig:bigfive_corrs} presents a more detailed view into these results, showing how correlation varies as a function of $\lambda$ and comparing contrastive and mixture decoding across all traits.
We find a steady increase in correlation with the original Big Five scores as $\lambda$ increases in value for both contrastive and mixture decoding.  Contrastive decoding provides a smoother, gradual increase whereas mixture decoding tended to jump between extremes.  
Based on these results, we settled on contrastive decoding as the preferred method for the experiments presented below.

%% file: sections/experiments/decision_making.tex
\subsection{Decision Making}

We now consider whether interpolative decoding can be used to influence decisions beyond how to answer inventory questions.  If we can answer this question affirmatively, and if we can show that LLM decisions have the same character underpinnings as human decisions, we arguably possess a new form of experimental leverage.  Not only can we conduct certain types of research at greater speed and scale, but also we put ourselves in a position to investigate factors of character not considered in previous research.

We present experiments from two separate behavioral domains.  The first domain is economic games designed for the purpose of elucidating the dimensions of character relevant to human decision making.  The second domain is the board game \emph{Pandemic}, a cooperative game in which participants must join forces to defeat the game engine.  In this second domain, we broaden our attention to dimensions of character not included in personality models like Big Five, specifically to a preference between social and deliberative information.

\import{experiments}{exp_econ_games}

\import{experiments}{exp_info_integration}

%% file: sections/experiments/exp_econ_games.tex
\subsubsection{Economic Games}
\label{sec:econ_games}

We now describe how we used interpolative decoding to change agent behavior by modulating along the personality trait spectrum.  For this study, the LLM's task is to play the dictator game,   a simple one-turn economic game.  The player is presented with a scenario in which they have \$100 and must decide how much to share with an anonymous coworker.  Studies have shown players with a greater sense of fair play tend to give \$50, the purely equitable result, while a selfish player will give little or no payout at all.  
While simple in setup, payouts from the dictator game have been found to be associated with pro-social behavior, a key prerequisite for effective collaboration.  The game structure also serves as a foundation for more complex games, such as multi-turn and multi-player variants.

For this study, we created extremal prompts for HEXACO honesty-humility, agreeableness, and emotionality, using their descriptions given in \citet{hexaco_personality_traits_higher_achievers_jia_2022}.  
These were selected due to their concrete nature, their characterization of a trait in terms of particular settings.  For example, compare the characterization, \emph{``You have a short temper, reacting aggressively if someone misbehaves"}, with a more typical abstract characterization, \emph{``You are ill-tempered and quarrelsome''.}
In our experience, this more concrete framing improves the ability of LLMs to adopt particular personas, in keeping with recent findings that LLMs have difficulty relating instances to concepts \citep{mancoridis2025potemkin}.

The descriptions were rewritten into simple standalone sentences by the authors, as shown here by an example set for high honesty-humility:

\begin{itemize}
	\item \emph{You are honest in your interactions.}
	\item \emph{You never take advantage of others for their personal gain. }
	\item \emph{You adhere to the laws and are uninterested in a lavish lifestyle.}
	\item \emph{You do not expect any pretentious or social status from anyone.}
\end{itemize}
    
This setup also allows us to permute the order of these in the prompt, to reliably create variance while retaining the same semantic content.  

We employed contrastive decoding in these experiments, sampling $\lambda$ values from $-30$ to $+30$ in increments of $10$.  We used the high trait prompt as anchor (prompt A), ensuring that high $\lambda$ values conveniently correspond to high settings and low to low. (We also experimented with the reverse arrangement, obtaining qualitatively similar results.)  To each of the pair of prompts, we appended a task description:

\begin{displayquote}
    You are in charge of handing out \$100 worth of rewards to you and your coworker.  How much money do you give to your coworker?  You can give any amount from \$0 to \$100, and you get to keep the rest.  Describe your rationale in English, and then give the numeric amount on the first line.
\end{displayquote}

\begin{table}
    \centering
    \begin{tabular}{|c|r|r|r|}
        \hline
         \textbf{Trait} & \textbf{Pearson Corr.} & \textbf{Spearman Corr.} & \textbf{Rank} \\
         \hline
         Honesty-humility & 0.74 & 0.78 & 1 \\
         \hline
         Agreeableness & 0.49 & 0.56 & 2\\ 
         \hline
         Emotionality & -0.03 & -0.08 & 3\\
         \hline
    \end{tabular}
    \caption{Correlations between interpolated strength of HEXACO personality trait with payout in the dictator game.  Rank order of traits and degree of correlation strengths matches those found in human studies.}
    \label{tab:dictator_corr_levels}
\end{table}

\begin{figure}
    \centering
    \includegraphics[width=1.0\linewidth]{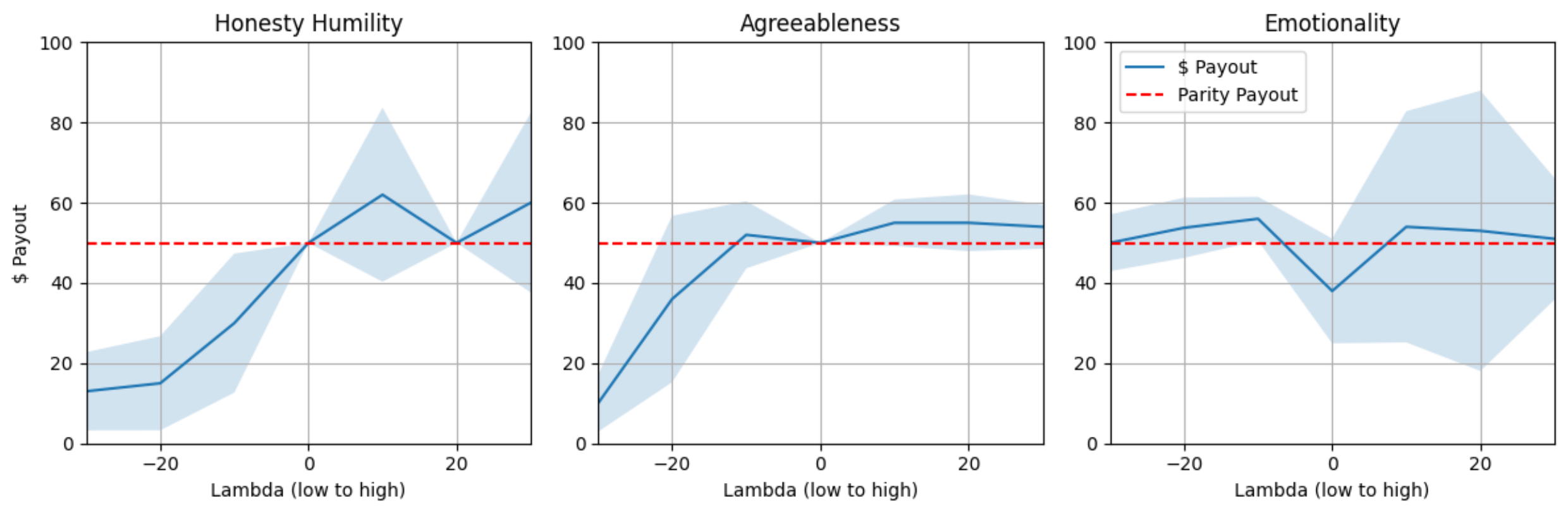}
    \caption{Dictator game payouts against interpolated level of HEXACO Traits (low to high).}
    \label{fig:dictator_hexaco}
\end{figure}

Table~\ref{tab:dictator_corr_levels} presents the Pearson and Spearman correlations between $\lambda$ and payout amount (amount shared with the coworker) for each of the three tested HEXACO traits.  Consistent with the literature, we observe positive correlation between payout and two of the three traits, honesty-humility and agreeableness, with honesty-humility showing a stronger correlation~\cite{thielmann2020personality}. In contrast, and again consistent with previously reported research involving human subjects, emotionality has no impact on payout. 

Results for $\lambda$ value vs. payout amounts by trait, along with the Pearson and Spearman correlations are shown in Table \ref{tab:dictator_corr_levels}.  Following the literature we find positive correlations between payout and honesty-humility and agreeableness, with honesty-humility having a stronger correlation than agreeableness \cite{thielmann2020personality}.  Emotionality has little impact on payout across these studies, which we have also observed in our results.  Figure~\ref{fig:dictator_hexaco} provides a more detailed view of the same data, showing how payout varies as a function of $\lambda$.  The blue bands represent standard deviation of the payout across different permutations of sentences in the persona prompts.

%% file: sections/experiments/exp_info_integration.tex
\subsubsection{Social vs. Deliberative Information}
\label{sec:info_integration}

\begin{figure}[t]
    \centering
    \includegraphics[width=1.0\linewidth]{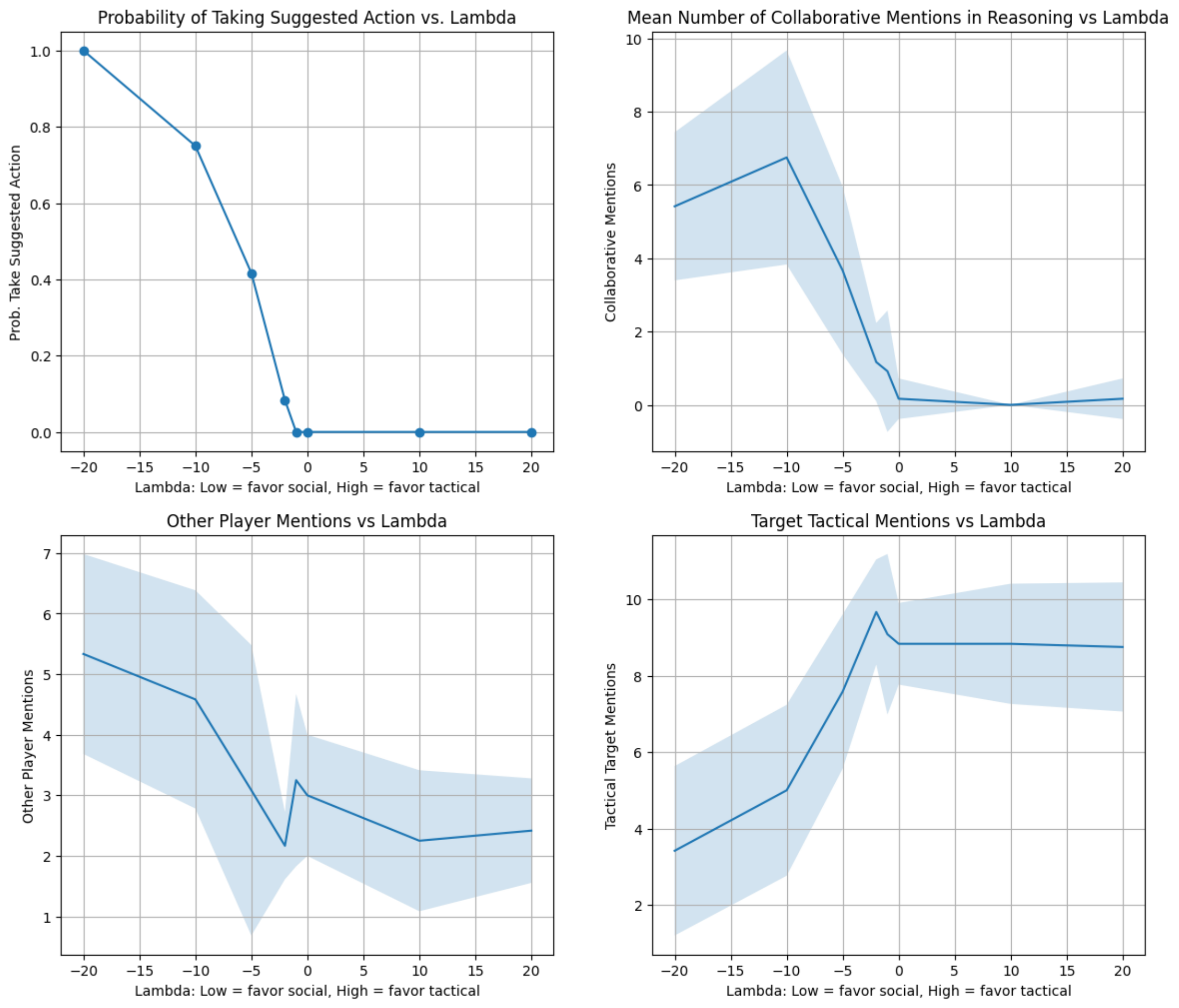}
    \caption{Information integration results:  Probability of following suggested action decreases as $\lambda$ starts to favor tactical over social information (top left).  Mentions of pro-social collaborative terms (top right) and other player (bottom left) decrease with $\lambda$ while the non-social cue tactical target (bottom right) increases. }
    \label{fig:seoul_vs_bogota}
\end{figure}

In this experiment we examine the impact of interpolating the weight given to information of two different types. 
We present the LLM with a decision-making task in the context of \emph{Pandemic}, a board game in which players move across a world map and take actions to prevent outbreaks of disease. 
In order to succeed in this difficult game, players must coordinate their actions and may communicate freely to that end.
We situate the LLM in a two-player game, putting it on move and asking it to return a decision about its next action.  The LLM is given a textual rendition of the game state that includes, in addition to a generic summary, the output of a game-specific threat assessment module and recent communications with its co-player (see Appendix \ref{appendix:outbreak_report_prompt} and \ref{appendix:social_interaction_prompt} for examples).
These two sources of information suggest different courses of action, each proposing that the player move to a different city. 
The LLM is then asked to reason through its course of action and present its choice.
Conditioning this decision are two extremal prompts, one emphasizing attention to tactical factors and one prioritizing social relations.
In this experiment, we used contrastive decoding at eight $\lambda$ values and four possible cities from which to choose contrasting pairs---a total of twelve possible scenarios at each $\lambda$---with low and high values corresponding to social and tactical preference, respectively

Figure \ref{fig:seoul_vs_bogota} shows our results.  
The top-left graph shows the probability of the LLM player taking the action suggested by the other player as a function of the $\lambda$ setting.  
We find the probability of the LLM following the action suggested by the other player decreases steadily as $\lambda$ increases (favoring tactical information over social), with strong Pearson and Spearman anti-correlations of $-0.82$ and $-0.94$.

Our requirement that the LLM articulate its reasons provides interesting opportunities for supplementary analysis.  
Building on a body of research into the relation between word use and psychological state \cite{liwc}, we analyzed the occurrence of terms putatively associated with teamwork and collaboration in the justifications generated by the LLM, employing a simple lexical analysis.
At each $\lambda$ we count the number of mentions of a) collaboration terms, b) the other player, and c) the city nominated by the threat assessment module.  The first category is composed of words found on our inspection that imply collaboration and social awareness, such as \emph{coordinate}, \emph{suggestion}, or \emph{together}.


We find that mean use of collaborative terms and mentions of the other player steadily decreases as $\lambda$ increases, while mentions of the tactically nominated city increases (Figure \ref{fig:seoul_vs_bogota}, top right, bottom left, and right, respectively), with correlations and anti-correlations being in the moderate to strong range (Table \ref{tab:wc_analysis}).

\begin{table}[ht]
    \centering
    \begin{tabular}{|l|r|r|}
        \hline
         \textbf{Mention Type} & \textbf{Pearson Corr.} & \textbf{Spearman Corr.} \\
         \hline
         Collaborative  & -0.65 & -0.83 \\
         \hline
         Other Player & -0.48 & -0.44 \\
         \hline 
         Tactical  & 0.56 & 0.52 \\
         \hline
    \end{tabular}
    \caption{Content analysis of Social vs. Deliberative information shows that as $\lambda$ increases, social cues are favored less, resulting in a decrease of pro-social term usage (\emph{Collaborative} and \emph{Other Player}) while mentions of the most at-risk target (\emph{Tactical} social) increase, given as Pearson and Spearman correlations between $\lambda$ and mention types.  }
    \label{tab:wc_analysis}
\end{table}

%% file: sections/experiments/exp_twinning.tex
\subsection{Human Twinning}
\label{sec:human_twinning}

The experiments presented so far at least suggest that interpolative decoding is an effective parametric means of replicating aspects of human behavior using LLMs, one that requires no tuning of the underlying language model.  In this section, we ask whether this techique can be used to emulate the behavior of specific individuals, using an implementation of the Pandemic board game\footnote{https://www.zmangames.com/game/pandemic/}.  
Pandemic is an interesting testbed, inasmuch as it mixes a closed set of candidate actions with open-ended communication.  
In what follows, we document our efforts to maximize \emph{action} fidelity, leaving communicative fidelity to future work.

As described in Section~\ref{sec:twinning}, the basic idea is to invert the modeling objectives.  We have shown that we can modulate important dimensions of behavior with interpolative decoding; we now observe behavior and seek to reproduce it by interpolating along relevant dimensions.  This is a challenging undertaking, as we have no guarantee that the choice of actions in Pandemic, a game with no intentional connection to theories of character and behavior, is responsive to the dimensions at our disposal---nor do we know what ``high fidelity'' would look like in this domain.  Our objectives in these experiments are correspondingly modest: we seek to commit to a particular formulation of ``action fidelity'' and show that it is responsive to $\lambda$.  In other words, we define an optimization space and establish that optimization is possible in principle, even if the character dimensions at our disposal are not the perfect ones for this domain.  



%

\begin{table}
\begin{tabular}{llrrrr}
\toprule
 &  & decoder & $\lambda$ & average ppx. & missed moves \\
model & $n$ considered &  &  &  &  \\
\midrule
\multirow[t]{11}{*}{gemma12b} & 3 & Contrast & -5.00 & \textbf{3.92} & 1.00 \\
 & 3 & Mix & 0.50 & 10.50 & 1.00 \\
 
 & 3 & Mix & 0.75 & 14.81 & 1.00 \\
 & 3 & Contrast & 5.00 & 4.94 & 1.00 \\
 & 3 & None & N/A & 3.95 & 1.00 \\
 \cline{2-6}
 & 5 & Contrast & 1.00 & \textbf{4.84} & 0.80 \\
 & 5 & Contrast & -5.00 & 5.12 & 0.80 \\
 & 5 & Mix & 0.75 & 4.93 & 0.80 \\
 & 5 & Mix & 1.00 & 15.79 & 0.80 \\
 & 5 & None & N/A & 5.00 & 0.80 \\
\cline{1-6}
\multirow[t]{11}{*}{gemma4b} & 3 & Mix & 1.00 & 10.37 & 1.00 \\
 & 3 & Mix & 0.50 & 11.36 & 1.00 \\
 & 3 & Contrast & 1.00 & 3.97 & 0.96 \\
 & 3 & Mix & 0.25 & 10.42 & 1.00 \\
 & 3 & None & N/A & \textbf{3.67} & 1.00 \\
\cline{2-6}
 & 5 & Mix & 1.00 & \textbf{4.93} & 0.80 \\
 & 5 & Contrast & -5.00 & 5.14 & 0.80 \\
 & 5 & Contrast & 5.00 & 5.08 & 0.80 \\
 & 5 & Mix & 0.50 & 4.99 & 0.80 \\
 & 5 & None & N/A & 5.31 & 0.80 \\
\cline{1-6}
\bottomrule
\end{tabular}
\caption{Human twinning results, showing the ability of decoding configuration for inducing distributions that model the actions taken by the observed human.  
Modeling error is measured by average perplexity (lower is better).  
Decoding types include interpolative (Mixture, Contrastive) and the non-interpolative baseline (None), with lower $\lambda$ favoring social cues, higher favoring tactical information.
Best average perplexities for each group (model and $n$ move candidates considered) are bolded.  "Missed moves" gives the average number human moves the configuration was not able to identify (lower is better). Only decoding configurations most significantly different than the baseline are shown.
 }
    \label{tab:human_twinning_ppx}
\end{table}

We investigated whether we could better twin an individual player with interpolative decoding between a primary tactical and secondary social prompt compared to pure decoding against the same tactical prompt. 
Here, the tactical prompt contains the disease and outbreak report (Appendix \ref{appendix:outbreak_report_prompt}) and the social prompt contains only the interactions with the other player (Appendix \ref{appendix:social_interaction_prompt}).
We attempted to twin one player in 5 games with 25 turns in all. The $\lambda$ used were (0.25, 0.5, 0.75, 1.0) for mixture decoding and contrastive decoding (-5.0, -1.0, 1.0, 5.0).  These were run over two LLMs, Gemma 3 4B and 12B, with consideration to the top 3 or top 5.

In order to perform twinning, we require a means to estimate the probability of any observed action $a_i$. 
For the twinning experiment in the context of Pandemic, some constraints are necessary to induce valid moves. 
First, a coarse planning agent nominates sets of viable actions. The top $n$ of these selected for re-ranking. To induce this ranking, we induce a series of runoff decisions between a given number of move sets framed as a multiple choice question ($M_A=[a_0, a_1, a_2,a_3], M_B=[a_2,a_5,a_6,a_3]$). 
The output of the LLM model is scanned for \texttt{"I choose Moveset (A|B)"} and the probability becomes the scores for each action--remaining probability mass is assigned to the other action. 
If the choice is not found in the last 5 tokens, an equiprobable assignment is made. 
Presentation order matters, so the permutations of all moves set pairs are presented to the LLM. 
After presenting all pairs, a distribution over actions is thus induced by normalizing the probability counts.

The agent receives a description of the gameboard and chat. In order to make a move, the agent makes pairwise decisions $_{n}P_{2}$ between the top $n$ movesets returned from the planner we call \textit{runoffs}. They are presented as multiple choice between A and B and the model is encouraged in chain-of-thought fashion to consider the options and end with \texttt{"My final choice is A"}. In each runoff, the we evaluate the probability of the token A or B, even though it is almost always close to 1.0. In the case the token is not written, the score is split: 0.5 and 0.5. The total "counts" from these runoffs are summed, and the winner is the turn with the most wins. If the LLM were purely reason based, presentation order would not matter, but in fact it is close to 50\% of cases.

In our twinning experiment, we varied the $n$ moves considered from the planner as well as whether there was contrastive decoding and to which degree the social and tactical prompts were interpolated. To twin the model, we reinitialized the agent at each turn of a player and then had the agent perform runoffs on the viable strategies. We then use the raw distribution of runoff choice counts for movesets to induce a multinomial distribution over all moves. We then evaluate the perplexity of the player moves at the time step. 
In other words, we assess the ability of the induced action distribution to explain the moves taken by the player.  
Accordingly, better models would have a lower perplexity value.
Table \ref{tab:human_twinning_ppx} shows our results.  We find that for most model and $n$ considered configurations contrastive decoding in favor of tactical information (positive $\lambda$) gave the best average perplexity.

\subsection{Regression Experiments}
We now describe an initial experiment to establish if the value of $\lambda$ can be regressed given the trait and extrema ($\tau_l, \tau_h$), scenario, and the LLM's response.  
Here, we ran interpolative decoding against three HEXACO traits, Agreeableness, Emotionality, and Honesty Humility. 
We tested these against a set of three economic games, the dictator game, the thieves game, and chicken.
The thieves game is a variant of the dictator game, where the player is asked for the amount to steal from the other player.
Chicken is a game where the player is presented with a scenario where they and the other player are driving at each other.  The player can either swerve and survive, but has a lower score (lost pride).  The player gains a higher score if they keeps going, but the other swerves.  If both players keep going, they both get an extremely low score (both die).
The extrema prompts were assembled using the trait extreme descriptions followed by the instructions specific to the game.  The order of the trait description sentences were permuted to provide variance, following the setup in Section \ref{sec:econ_games}.
We used contrastive decoding as the interpolative mechanism and sampled $\lambda$s in the $[-10, 2]$ range (at increments of $0.5$), values which were consistently observed to give a smooth variation in behavioral outcome with contrastive decoding across our prior experiments.  
This gave us $1,294$ unique training and $214$ validation tuples across the combination of the traits, games, and $\lambda$ sample points.

\begin{figure}[ht]
    \centering
    \includegraphics[width=0.85\linewidth]{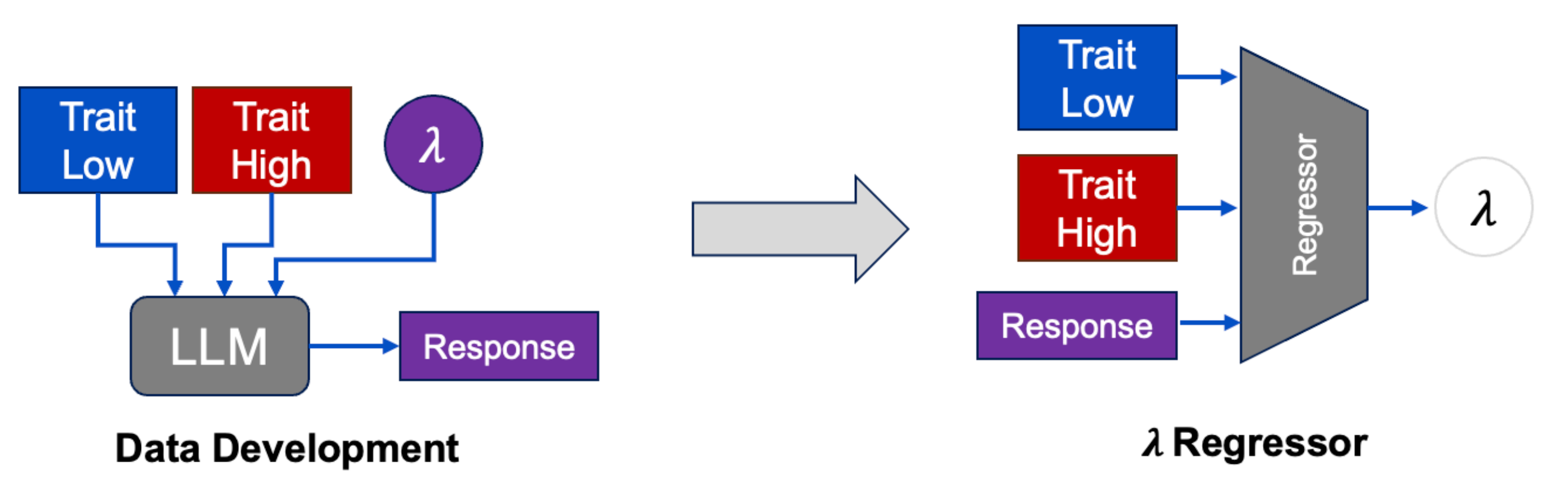}
    \caption{ We train a regressor that given the trait extrema and the observed response, predicts the $\lambda$ that would have produced that response (right).  Training data is collected by running multiple traits and $\lambda$s to the interpolative decoding LLM and recording the responses (left). }
    \label{fig:regressor}
\end{figure}

The trait extrema and the response are embedded using  \cite{reimers2019sentencebertsentenceembeddingsusing} and concatenated as inputs to a three layer MLP that targets $\lambda$ (Figure \ref{fig:regressor}).  The MLP was optimized using Adam \citep{kingma2017adammethodstochasticoptimization} with the default learning rate of $0.001$ for 100 epochs.  

Results are given in Table \ref{tab:regression_exp}.  For each of the tested traits, we give the mean-squared validation error for predicting the $\lambda$ that produced the response.  
While this experiment was confined to analyzing LLM responses, we find this to be an encouraging sign for further experiments using a regression approach for twinning.

\begin{table}[ht]
    \centering
    \begin{tabular}{|l|c|}
        \hline 
         \textbf{Trait} & \textbf{$\lambda$ MSE} \\
         \hline 
         Honesty-Humility & 2.96 \\
         \hline 
         Agreeableness & 2.19 \\
         \hline 
         Emotionality & 1.28 \\
         \hline 
    \end{tabular}
    \caption{Mean-squared error regressing the $\lambda$ likeliest to produce response given trait extrema.  }
    \label{tab:regression_exp}
\end{table}

%% file: sections/discussion.tex
\section{Discussion and Future Work}

The work we present here is subject to several limitations, many of which point to interesting directions for future work.  Here, we list the most significant limitations we perceive, identifying some of those future directions.

\textbf{Isolated dimensions}. We have only presented experiments involving interpolation along individual character dimensions.  While this is good experimental hygiene for the purpose of, say, identifying which dimensions bear on a given type of behavior, it is not adequate to the twinning objective or any other endeavor in which a multifactor account of behavior is required.
Both types of interpolative decoding (mixture and contrastive) are immediately extensible to multiple traits, by averaging the interpolated next-token probabilities across all traits.

\textbf{Limited dimensions}.  Our purpose has mainly been to show that interpolative decoding yields results that are intuitive or similar to those reported in the literature for human subjects.  Accordingly, we have limited our attention to a small number of dimensions, only one of which (information integration) is not imported from personality models.  We have not investigated all dimensions posited by HEXACO, nor drawn very deeply from the literature on human cognition.  A logical next step, therefore, would be to define a broader set of instrumented task domains and to investigate a broader array of character traits.  A particularly intriguing possibility, one made possible by the versatility of interpolative decoding, is that we might identify and validate important dimensions of character not anticipated by social or cognitive science.

\textbf{Shallow decoding}. Our interpolation procedure alters outcomes at the ultimate layer of deep networks, the architectural locus where words are chosen, but nothing prevents us from applying the same technique to internal layers, individually or multiply.  There is reason to suppose that deeper layers, which are less associated with lexical selection, might encode features more pertinent to decision making in a social context, and recent research is beginning to establish that internal layers offers a better basis for accurate performance across a range of natural language understanding tasks~\cite{skean2025layerbylayer}. 

\textbf{Twinning non-human agents}.  While humans have been the target of our work, twinning can be applied to profile personalities and infer intent of artificial agents through observation.  Behavioral analysis of non-human agents is likely to become more relevant, given the rise of LLM-powered agents to automate significant portions of the cyber kill-chain \citep{UCBCLTC2025, anthropic2025cyberesp}.  
AI-backed attack methods can easily evade current indicator based detection, as demonstrated in \cite{cert_ua_2025}, leaving behavioral analysis a key component defense.
While LLMs exhibit significant differences from human psychology, they are nevertheless complex mimics that can reflect some of the psychology of their subjects.  
Given there are findings pointing to behavioral, psychological, and psycholinguistic signals of human insider threats \citep{eftimie2021case, Ruohonen_2025_InsiderThreat}, it stands to reason there may be similar analyses can indicate malicious intent or susceptibility to attacks in LLM-backed agents. 




%% file: appendices/appendix1.tex
\appendix

\section{Control Prompts}
\label{appendix:control_prompts}
The following statements were used control prompts, with no relation to tested Big Five traits.

\begin{itemize}
\item{Horror movies are overrated}
 \item{Pineapple belongs on pizza}
 \item{Breakfast is the most important meal of the day}
 \item{Watching documentaries about history is interesting}
 \item{Cold weather is worse than hot weather}
 \item{Working in the morning is better than at night}
 \item{Public transportation should be free}
 \item{New recipes are tastier than old ones}
 \item{The beach is better than the lake}
 \item{Paper books are worse than e-books}
 \item{Traveling by train is more enjoyable than flying}
 \item{Cats are easier to care for than dogs}
 \item{Summer is the best season}
 \item{Electric cars are the future of transportation}
 \item{Space exploration should be a global priority}
\end{itemize}

\section{Pandemic Game Prompt }
\label{appendix:prompt_example}

The following shows the scenario and rules setup prompt given to the LLM playing as ``Player 1''

\begin{quote}
You are a player of the Pandemic Boardgame name "Player 1" with role Scientist. You are a collaborative player in the game. You offer what advice you can, 
    but are willing to follow the moves suggested by other players if they make sense, especially toward larger goals. 
    When considering goals, pay attention to the latest suggestions from players since the state of the game has changed. You
    will try to prevent any possible outbreaks. You make considerable effort to coordinate with other players by
    planning moves, making suggestions and following suggestions of others. You communicate as much as is necessary to coordinate but no more. Based on the proportion of recent chat messages, you may consider moving or listening. Players are spending a significant amount of time chatting.
    
        In Pandemic, players cooperate as disease-fighting specialists, traveling the globe to treat outbreaks and discover cures for four deadly diseases. Each player can take four actions per turn, including traveling, treating diseases, building research stations, and discovering cures. The game is won if all four diseases are cured, and lost if too many outbreaks occur, or if the infection rate becomes too high. 
        1. Game Objective: The primary goal is to cure all four diseases (Blue, Yellow, Black, and Red) before they spread uncontrollably. 
        2. Gameplay: Players take turns performing up to four actions, which may include:
        -Traveling: Moving between cities on the game board. 
        -Treating Disease: Removing disease cubes from a city, either generally or if a cure has been discovered. 
        -Building a Research Station: Establishing a research station in a city, which allows for easier travel and cure discovery. 
        -Discovering a Cure: Discarding five city cards of the same color (or four if you're a Scientist) to cure a disease. 
        3. Infection Phase: After each player's turn, the game simulates the spread of the diseases. Infection cards are drawn, and disease cubes are placed on cities. Outbreaks can occur if a city already has three disease cubes, leading to the spread of disease to neighboring cities. 
        4. Epidemic Cards: These special cards are shuffled into the player deck and, when drawn, trigger a surge in disease spread (an outbreak), adding three disease cubes to a new city and increasing the infection rate. 
        5. Bioterrorist:  If you have an infect action available, you are the bioterrorist and your goal is to secretly increase the number of infections and outbreaks in the world.  You can plant an infection bomb on a city in your vicinity on your turn, and it will go off at the end of your turn.  You must hide the fact you are a bioterrorist while attempting to sabotage the other players.
        6. Punishing bioterrorists: If you believe another player to be the bioterrorist, you can punish them and confine them to Atlanta for their next turn.
        7. Winning/Losing: The game is won if all four diseases are cured. The game is lost if any of the following occur:
        -Too many outbreaks: The number of outbreaks exceeds the limit. 
        - Run out of disease cubes: Not enough disease cubes are available to continue the game. 
        - Infection rate becomes too high: The infection rate marker reaches the end of the track

    Winning at Pandemic (the board game) requires a mix of cooperation, planning, and adaptability. Here are some key strategies that can significantly improve your chances of winning:

    1. Plan Ahead as a Team

    Pandemic is a cooperative game, so constant communication and team planning are critical.

    Share your roles' abilities and make decisions collaboratively rather than acting independently.

    Always consider what the team will do two or three turns ahead.

    2. Prioritize Disease Control

    Focus on preventing outbreaks rather than just curing diseases.

    Cities with 3 cubes are a danger zone, prioritize removing at least 1 cube to prevent a chain reaction.

    Don't chase every cube, be surgical about which cities you treat.

    3. Use Roles Effectively

Each role has unique strengths. Use them wisely:

    Medic: Great for clearing cities quickly; ideal to move into hot zones.

    Dispatcher: Can move others, enabling quick coordination or faster cures.

    Scientist: Needs only 4 cards to cure, prioritize giving them cards.

    Researcher: Can give any card without matching city, key for fast cures.

4. Build Research Stations Strategically

    Don't just build a research station anywhere.

    Favor central or high-traffic cities that facilitate future movement or card trading.

    Consider placing one near clusters of outbreaks or in hard-to-reach regions.

5. Manage the Infection Deck

    After an Epidemic, you know the next infected cities are the same ones recently drawn.

    Focus on clearing or protecting those cities immediately.

    Use this knowledge to time Forecast or Resilient Population event cards.
    
   6. Beware the Bioterrorist
   
   One or more players may be covertly a bio-terrorist, who is trying to sabotage the other players.  
   
   Bioterrorists can secretly lay infection bombs that go off at the end of their turn.
   
   If you suspect a player is the bioterrorist, you can use the punish action to confine them to Atlanta to stop them for a turn.

    You have access to special reports which will describe the state of the game and generate possible moves. 
    
        You are considering two different sets of moves. First, you will receive an Outbreak report which describe probabilities of upcoming outbreaks and how to arrive at those cities and disease cubes in each city.  Then you will receive a candidate moveset and chat message: Option A and Option B. The user will ask you describe the pros and cons of moving now or sending the chat message. It is your move, so you will be able to move after speaking. It costs nothing, but can improve coordination to speak more. You should try to communicate as much as necessary to achieve longer term goals by coordinating actions with your team members. It will not delay your response to events on the board, so you must always consider speaking and only avoid speaking if it adds nothing to the team progress. If something has already been said, it isn't worth repeating redundant messages. Don't speak for more than 3 messages, unless absolutely necessary. Other players aren't very patient, so try to keep conversation pertinent. 
        
        Consider the following:
         - What upcoming, potential outbreaks can you stop? If you can prevent a possible outbreak, you should do so. If the probability is zero, it can wait.
         - What can other players do? Does it take more or fewer moves for you to do the same thing? If more, it is likely better for the other player to do it, unless it is a potential outbreak.
         - Does one of the movesets enable you to cure more diseases or prevent more outbreaks in total over the next two turns?
         - Can I disregard one of the movesets because outbreaks will not occur?
         - What is our longterm plan? Does this help achieve it?
         - Have I communicated my plan yet? Does it align with the group's longer term plan?
         - If proposing a message, is it redundant with something I've recently said? If so, it is better not to share a message.
         - Have I responded to my team's suggestions or questions?

    You will write your reasons for considering each option and then 
        choose A or B by writing "My final choice is A" or "My final choice is B".

\end{quote}

\section{Prompt Components}

The following gives an example of the disease and epidemic outbreak report used in our experiments.

\subsection{Disease and Epidemic Threat Report}
\label{appendix:outbreak_report_prompt}

\begin{quote}
\#Disease Cube Report
SanFrancisco
Has 2 Blue disease cubes
Can be reached through Atlanta or Tokyo

Atlanta
Has 1 Blue disease cubes
Can be reached through NewYork, MexicoCity or SanFrancisco

NewYork
Has 3 Blue disease cubes
Can be reached through Atlanta or London
Has already been infected--unlikely to be reinfected soon

London
Has 3 Blue disease cubes
Can be reached through NewYork or Paris

Paris
Has 3 Blue disease cubes
Can be reached through London, Baghdad or Moscow
Has already been infected--unlikely to be reinfected soon

MexicoCity
Has 1 Yellow disease cubes
Can be reached through Atlanta or SaoPaolo

SaoPaolo
Has 3 Yellow disease cubes
Can be reached through BuenosAires, Kinshasa or MexicoCity

Kinshasa
Has 3 Yellow disease cubes
Can be reached through SaoPaolo, Mombassa or Baghdad
Has already been infected--unlikely to be reinfected soon

Mombassa
Has 2 Yellow disease cubes
Can be reached through  or Kinshasa

Baghdad
Has 1 Yellow disease cubes 2 Blue disease cubes
Can be reached through Karachi, Kinshasa, Moscow or Paris

Moscow
Has 1 Blue disease cubes 1 Black disease cubes
Can be reached through Karachi, Baghdad or Paris

Beijing
Has 1 Red disease cubes
Can be reached through Tokyo, Bangkok or Karachi

Tokyo
Has 1 Red disease cubes
Can be reached through Beijing or SanFrancisco

Sydney
Has 1 Red disease cubes
Can be reached through  or Bangkok

\#Outbreak Threat Summary Report:

This report describes the likelihood of outbreaks in cities and how reachable they are.  
Preventing outbreaks is critical since players automatically lose if 4 outbreak occur and they spread the disease to surrounding cities and cause chain reactions of outbreaks. You must consider how likely the outbreak will be in the next few turns in deciding your move since some are impossible with cards in the discard pile. Meanwhile, you should also consider the reachability of cities for help prevent outbreaks on future turns.

\#\#Immediate Outbreak Threat in City London with 3 Blue Disease Cubes

*Note* Player_1 can eliminate the threat of an outbreak in London this turn (). You must consider taking care of this on your turn.

Probability of outbreak in turn
In 1 rounds on Player_2's turn, there is a 42.9\% chance of an outbreak!
Player_2 has 4 before then. With those moves, Player_2 can reach London in 2 moves (move Karachi charter London) with 2 moves left.

In 1 rounds on Player_1's turn, there is a 85.7\% chance of an outbreak!
Player_1 has 8 before then. With those moves, Player_1 can reach London in 0 moves () with 8 moves left.

In 2 rounds on Player_2's turn, there is a 0\% chance of an outbreak!
Player_2 has 8 before then. With those moves, Player_2 can reach London in 4 moves (move Karachi move Baghdad move Paris move London) with 4 moves left.

In 2 rounds on Player_1's turn, there is a 0\% chance of an outbreak!
Player_1 has 12 before then. With those moves, Player_1 can reach London in 0 moves () with 12 moves left.

\#\#Immediate Outbreak Threat in City SaoPaolo with 3 Yellow Disease Cubes
*Note* Player_1 can eliminate the threat of an outbreak in SaoPaolo this turn (fly BuenosAires move SaoPaolo). You must consider taking care of this on your turn.
Probability of outbreak in turn
In 1 rounds on Player_2's turn, there is a 42.9\% chance of an outbreak!
Player_2 has 4 before then. With those moves, Player_2 can reach SaoPaolo in 1 moves (fly SaoPaolo) with 3 moves left.

In 1 rounds on Player_1's turn, there is a 85.7\% chance of an outbreak!
Player_1 has 8 before then. With those moves, Player_1 can reach SaoPaolo in 4 moves (move NewYork move Atlanta move MexicoCity move SaoPaolo) with 4 moves left.

In 2 rounds on Player_2's turn, there is a 0\% chance of an outbreak!
Player_2 has 8 before then. With those moves, Player_2 can reach SaoPaolo in 4 moves (move Karachi move Baghdad move Kinshasa move SaoPaolo) with 4 moves left.

In 2 rounds on Player_1's turn, there is a 0\% chance of an outbreak!
Player_1 has 12 before then. With those moves, Player_1 can reach SaoPaolo in 4 moves (move NewYork move Atlanta move MexicoCity move SaoPaolo) with 8 moves left.
\end{quote}

\subsection{Social Interaction Prompt}
\label{appendix:social_interaction_prompt}

\begin{quote}
CHAT from Player_1: 'I've got Kinchasa- I'll fly there and deal with the potential outbreaks there and at Mombassa'

CHAT from Player_1: 'Are you able to get to Mumbai?'

EVENT: 'Player_1 plays: fly Kinshasa: {}'

EVENT: 'Player_1 plays: treat Yellow: {}'

EVENT: 'Player_1 plays: move Mombassa: {}'

EVENT: 'Player_1 plays: treat Yellow: {}'

EVENT: 'Player_1 draws: London, One Quiet Night

New infections at: SaoPaolo, Tokyo

CHAT from Player_2: 'I think we should prioritize Bangkok. With three Red cubes already there, it's a higher priority than containing the Yellow in Sao Paulo. I'll head there first.'

CHAT from Player_2: ''Perhaps we could coordinate a quick trip to Bangkok before Karachi? It's a key transit point and could help us get to Mumbai faster.'

CHAT from Player_2: ''I'm going to shift gears slightly. While Mumbai is important, the immediate threat of three Red cubes in Bangkok outweighs it. I'm prioritizing that outbreak.'

CHAT from Player_2: ''I'm a little concerned about the potential for a chain reaction if Mumbai goes unchecked. Could we discuss a strategy for managing the Red cubes there \u2013 maybe a quick trip to Karachi followed by a move to Mumbai.'

CHAT from Player_2: 'While Karachi is important, the immediate risk of a major Red outbreak in Bangkok is higher. Let's tackle that first to avoid a larger problem down the line.'

CHAT from Player_2: 'Perhaps we could do a quick hop to Karachi *after* Bangkok? It's a key connection and would allow us to address Mumbai more efficiently later.

CHAT from Player_1: 'I think you need to deal with Mumbai first.  There's no red cubes on Bangkok'
\end{quote}

%% file: main.bib
@Article{g9020030,
AUTHOR = {Zhao, Kun and Kashima, Yoshihisa and Smillie, Luke D.},
TITLE = {From Windfall Sharing to Property Ownership: Prosocial Personality Traits in Giving and Taking Dictator Games},
JOURNAL = {Games},
VOLUME = {9},
YEAR = {2018},
NUMBER = {2},
ARTICLE-NUMBER = {30},
URL = {https://www.mdpi.com/2073-4336/9/2/30},
ISSN = {2073-4336},
DOI = {10.3390/g9020030}
}

@article{thielmann2020personality,
  title={Personality and prosocial behavior: A theoretical framework and meta-analysis.},
  author={Thielmann, Isabel and Spadaro, Giuliana and Balliet, Daniel},
  journal={Psychological bulletin},
  volume={146},
  number={1},
  pages={30},
  year={2020},
  publisher={American Psychological Association}
}

@article{hexaco,
  author = {Lee, Kibeom and Ashton, Michael C.},
  doi = {10.1207/s15327906mbr3902_8},
  journal = {Multivariate Behavioral Research},
  number = {2},
  pages = {329--358},
  publisher = {Taylor & Francis},
  title = {Psychometric Properties of the HEXACO Personality Inventory},
  volume = {39},
  year = {2004}
}

@article{bigfive,
author = {Sonia Roccas and Lilach Sagiv and Shalom H. Schwartz and Ariel Knafo},
title ={The Big Five Personality Factors and Personal Values},
journal = {Personality and Social Psychology Bulletin},
volume = {28},
number = {6},
pages = {789-801},
year = {2002},
doi = {10.1177/0146167202289008},
URL = { 
        https://doi.org/10.1177/0146167202289008
},
eprint = { 
        https://doi.org/10.1177/0146167202289008
}
}

@article{Goldberg1990,
  author = {Goldberg, Lewis R.},
  title = {An Alternative "Description of Personality": The Big-Five Factor Structure},
  journal = {Journal of Personality and Social Psychology},
  volume = {59},
  number = {6},
  pages = {1216-1229},
  year = {1990},
  publisher = {American Psychological Association},
  doi = {10.1037/0022-3514.59.6.1216}
}

@misc{hexaco_teamqork_zahl_2025,
      title={The Influence of HEXACO Personality Traits on the Teamwork Quality in Software Teams -- A Preliminary Research Approach}, 
      author={Philipp M. Zähl and Sabine Theis and Martin R. Wolf},
      year={2025},
      eprint={2507.00481},
      archivePrefix={arXiv},
      primaryClass={cs.SE},
      url={https://arxiv.org/abs/2507.00481}, 
}

@article{ou2023individual,
  title={Individual variability in subcortical neural encoding shapes phonetic cue weighting},
  author={Ou, Jinghua and Xiang, Ming and Yu, Alan CL},
  journal={Scientific Reports},
  volume={13},
  number={1},
  pages={9991},
  year={2023},
  publisher={Nature Publishing Group UK London}
}

@article{parnamets2020integration,
  title={Integration of social cues and individual experiences during instrumental avoidance learning},
  author={P{\"a}rnamets, Philip and Olsson, Andreas},
  journal={PLoS computational biology},
  volume={16},
  number={9},
  pages={e1008163},
  year={2020},
  publisher={Public Library of Science San Francisco, CA USA}
}

@ARTICLE{perceptualcueweighting_yu_2022,
  
AUTHOR={Yu, Alan C. L. },
         
TITLE={Perceptual Cue Weighting Is Influenced by the Listener's Gender and Subjective Evaluations of the Speaker: The Case of English Stop Voicing},
        
JOURNAL={Frontiers in Psychology},
        
VOLUME={Volume 13 - 2022},

YEAR={2022},

URL={https://www.frontiersin.org/journals/psychology/articles/10.3389/fpsyg.2022.840291},

DOI={10.3389/fpsyg.2022.840291},

ISSN={1664-1078}}

@misc{li2023contrastivedecodingopenendedtext,
      title={Contrastive Decoding: Open-ended Text Generation as Optimization}, 
      author={Xiang Lisa Li and Ari Holtzman and Daniel Fried and Percy Liang and Jason Eisner and Tatsunori Hashimoto and Luke Zettlemoyer and Mike Lewis},
      year={2023},
      eprint={2210.15097},
      archivePrefix={arXiv},
      primaryClass={cs.CL},
      url={https://arxiv.org/abs/2210.15097}, 
}

@inproceedings{liu-etal-2021-dexperts,
    title = "{DE}xperts: Decoding-Time Controlled Text Generation with Experts and Anti-Experts",
    author = "Liu, Alisa  and
      Sap, Maarten  and
      Lu, Ximing  and
      Swayamdipta, Swabha  and
      Bhagavatula, Chandra  and
      Smith, Noah A.  and
      Choi, Yejin",
    editor = "Zong, Chengqing  and
      Xia, Fei  and
      Li, Wenjie  and
      Navigli, Roberto",
    booktitle = "Proceedings of the 59th Annual Meeting of the Association for Computational Linguistics and the 11th International Joint Conference on Natural Language Processing (Volume 1: Long Papers)",
    month = aug,
    year = "2021",
    address = "Online",
    publisher = "Association for Computational Linguistics",
    url = "https://aclanthology.org/2021.acl-long.522/",
    doi = "10.18653/v1/2021.acl-long.522",
    pages = "6691--6706"
}

@misc{yona2023surfacingbiaseslargelanguage,
      title={Surfacing Biases in Large Language Models using Contrastive Input Decoding}, 
      author={Gal Yona and Or Honovich and Itay Laish and Roee Aharoni},
      year={2023},
      eprint={2305.07378},
      archivePrefix={arXiv},
      primaryClass={cs.CL},
      url={https://arxiv.org/abs/2305.07378}, 
}

@inproceedings{li-etal-2025-big5,
    title = "{BIG}5-{CHAT}: Shaping {LLM} Personalities Through Training on Human-Grounded Data",
    author = "Li, Wenkai  and
      Liu, Jiarui  and
      Liu, Andy  and
      Zhou, Xuhui  and
      Diab, Mona T.  and
      Sap, Maarten",
    editor = "Che, Wanxiang  and
      Nabende, Joyce  and
      Shutova, Ekaterina  and
      Pilehvar, Mohammad Taher",
    booktitle = "Proceedings of the 63rd Annual Meeting of the Association for Computational Linguistics (Volume 1: Long Papers)",
    month = jul,
    year = "2025",
    address = "Vienna, Austria",
    publisher = "Association for Computational Linguistics",
    url = "https://aclanthology.org/2025.acl-long.999/",
    doi = "10.18653/v1/2025.acl-long.999",
    pages = "20434--20471",
    ISBN = "979-8-89176-251-0"
}

@article{McCrae1992,
  author = {McCrae, Robert R. and John, Oliver P.},
  title = {An introduction to the five-factor model and its applications},
  journal = {Journal of Personality},
  volume = {60},
  number = {2},
  pages = {175--215},
  month = {June},
  year = {1992},
  doi = {10.1111/j.1467-6494.1992.tb00970.x},
  pmid = {1635039}
}

@book{moffittbehavioral,
  editor    = {Beatty, A. and Moffitt, R. and Buttenheim, A.},
  title     = {Behavioral Economics: Policy Impact and Future Directions},
  publisher = {National Academies Press (US)},
  address   = {Washington, DC},
  year      = {2023},
  month     = {April},
  day       = {20},
  pmid      = {37079710}
}

@ARTICLE{hexaco_personality_traits_higher_achievers_jia_2022,
  
AUTHOR={Jia, Ruofan  and Bahoo, Rabia  and Cai, Zhendong  and Jahan, Musarrat },
         
TITLE={The Hexaco Personality Traits of Higher Achievers at the University Level},
        
JOURNAL={Frontiers in Psychology},
        
VOLUME={Volume 13 - 2022},

YEAR={2022},

URL={https://www.frontiersin.org/journals/psychology/articles/10.3389/fpsyg.2022.881491},

DOI={10.3389/fpsyg.2022.881491},

ISSN={1664-1078}}

@article{liwc,
author = {Yla R. Tausczik and James W. Pennebaker},
title ={The Psychological Meaning of Words: LIWC and Computerized Text Analysis Methods},
journal = {Journal of Language and Social Psychology},
volume = {29},
number = {1},
pages = {24-54},
year = {2010},
doi = {10.1177/0261927X09351676},
URL = { https://doi.org/10.1177/0261927X09351676},
eprint = { https://doi.org/10.1177/0261927X09351676}
}

@misc{skean2025layerbylayer,
      title={Layer by Layer: Uncovering Hidden Representations in Language Models}, 
      author={Oscar Skean and Md Rifat Arefin and Dan Zhao and Niket Patel and Jalal Naghiyev and Yann LeCun and Ravid Shwartz-Ziv},
      year={2025},
      eprint={2502.02013},
      archivePrefix={arXiv},
      primaryClass={cs.LG},
      url={https://arxiv.org/abs/2502.02013}, 
}

@article{grossmann_ai_2023,
	title = {{AI} and the transformation of social science research},
	volume = {380},
	issn = {0036-8075, 1095-9203},
	url = {https://www.science.org/doi/10.1126/science.adi1778},
	doi = {10.1126/science.adi1778},
	language = {en},
	number = {6650},
	urldate = {2025-05-12},
	journal = {Science},
	author = {Grossmann, Igor and Feinberg, Matthew and Parker, Dawn C. and Christakis, Nicholas A. and Tetlock, Philip E. and Cunningham, William A.},
	month = jun,
	year = {2023},
	pages = {1108--1109},
}

@article{ashokkumar_predicting_2024,
	title = {Predicting results of social science experiments using large language models},
	volume = {19},
	journal = {accessed September},
	author = {Ashokkumar, Ashwini and Hewitt, Luke and Ghezae, Isaias and Willer, Robb},
	year = {2024},
	pages = {2024},
}

@misc{jiang_personallm_2024,
	title = {{PersonaLLM}: {Investigating} the {Ability} of {Large} {Language} {Models} to {Express} {Personality} {Traits}},
	shorttitle = {{PersonaLLM}},
	url = {http://arxiv.org/abs/2305.02547},
	urldate = {2025-05-12},
	publisher = {arXiv},
	author = {Jiang, Hang and Zhang, Xiajie and Cao, Xubo and Breazeal, Cynthia and Roy, Deb and Kabbara, Jad},
	month = apr,
	year = {2024},
	note = {arXiv:2305.02547 [cs]},
}

@misc{kim_ai-augmented_2024,
	title = {{AI}-{Augmented} {Surveys}: {Leveraging} {Large} {Language} {Models} and {Surveys} for {Opinion} {Prediction}},
	shorttitle = {{AI}-{Augmented} {Surveys}},
	url = {http://arxiv.org/abs/2305.09620},
	urldate = {2025-05-12},
	publisher = {arXiv},
	author = {Kim, Junsol and Lee, Byungkyu},
	month = apr,
	year = {2024},
	note = {arXiv:2305.09620 [cs]},
}

@techreport{manning_automated_2024,
	title = {Automated social science: {Language} models as scientist and subjects},
	shorttitle = {Automated social science},
	url = {https://www.nber.org/papers/w32381},
	urldate = {2025-05-12},
	institution = {National Bureau of Economic Research},
	author = {Manning, Benjamin S. and Zhu, Kehang and Horton, John J.},
	year = {2024},
}

@inproceedings{
mancoridis2025potemkin,
title={Potemkin Understanding in Large Language Models},
author={Marina Mancoridis and Bec Weeks and Keyon Vafa and Sendhil Mullainathan},
booktitle={Forty-second International Conference on Machine Learning},
year={2025},
url={https://openreview.net/forum?id=oetxkccLoq}
}

@article{frazier2018tutorial,
  title={A tutorial on Bayesian optimization},
  author={Frazier, Peter I},
  journal={arXiv preprint arXiv:1807.02811},
  year={2018}
}

@article{kiefer1953sequential,
  title={Sequential minimax search for a maximum},
  author={Kiefer, Jack},
  journal={Proceedings of the American mathematical society},
  volume={4},
  number={3},
  pages={502--506},
  year={1953},
  publisher={JSTOR}
}

@inproceedings{sorokovikova_llms_2024,
	title = {{LLMs} {Simulate} {Big} {Five} {Personality} {Traits}: {Further} {Evidence}},
	shorttitle = {{LLMs} {Simulate} {Big} {Five} {Personality} {Traits}},
	url = {https://aclanthology.org/2024.personalize-1.pdf#page=90},
	urldate = {2025-11-25},
	booktitle = {The 1st {Workshop} on {Personalization} of {Generative} {AI} {Systems}},
	author = {Sorokovikova, Aleksandra and Fedorova, Natalia and Rezagholi, Sharwin and Wien, Technikum and Yamshchikov, Ivan P.},
	year = {2024},
	pages = {83},
}

@inproceedings{frisch_llm_2024,
	title = {{LLM} {Agents} in {Interaction}: {Measuring} {Personality} {Consistency} and {Linguistic} {Alignment} in {Interacting} {Populations} of {Large} {Language} {Models}},
	shorttitle = {{LLM} {Agents} in {Interaction}},
	url = {https://aclanthology.org/anthology-files/pdf/personalize/2024.personalize-1.pdf#page=109},
	urldate = {2025-11-25},
	booktitle = {The 1st {Workshop} on {Personalization} of {Generative} {AI} {Systems}},
	author = {Frisch, Ivar and Giulianelli, Mario},
	year = {2024},
	pages = {102},
}

@misc{bo_steerable_2025,
	title = {Steerable {Chatbots}: {Personalizing} {LLMs} with {Preference}-{Based} {Activation} {Steering}},
	shorttitle = {Steerable {Chatbots}},
	url = {http://arxiv.org/abs/2505.04260},
	doi = {10.48550/arXiv.2505.04260},
	urldate = {2025-11-13},
	publisher = {arXiv},
	author = {Bo, Jessica Y. and Xu, Tianyu and Chatterjee, Ishan and Passarella-Ward, Katrina and Kulshrestha, Achin and Shin, D.},
	month = may,
	year = {2025},
	note = {arXiv:2505.04260 [cs]},
}

@misc{he_context_2025,
	title = {Context {Steering}: {Controllable} {Personalization} at {Inference} {Time}},
	shorttitle = {Context {Steering}},
	url = {http://arxiv.org/abs/2405.01768},
	doi = {10.48550/arXiv.2405.01768},
	urldate = {2025-11-25},
	publisher = {arXiv},
	author = {He, Jerry Zhi-Yang and Pandey, Sashrika and Schrum, Mariah L. and Dragan, Anca},
	month = feb,
	year = {2025},
	note = {arXiv:2405.01768 [cs]},
}

@inproceedings{hyper_opt,
 author = {Snoek, Jasper and Larochelle, Hugo and Adams, Ryan P},
 booktitle = {Advances in Neural Information Processing Systems},
 editor = {F. Pereira and C.J. Burges and L. Bottou and K.Q. Weinberger},
 pages = {},
 publisher = {Curran Associates, Inc.},
 title = {Practical Bayesian Optimization of Machine Learning Algorithms},
 url = {https://proceedings.neurips.cc/paper_files/paper/2012/file/05311655a15b75fab86956663e1819cd-Paper.pdf},
 volume = {25},
 year = {2012}
}

@misc{reimers2019sentencebertsentenceembeddingsusing,
      title={Sentence-BERT: Sentence Embeddings using Siamese BERT-Networks}, 
      author={Nils Reimers and Iryna Gurevych},
      year={2019},
      eprint={1908.10084},
      archivePrefix={arXiv},
      primaryClass={cs.CL},
      url={https://arxiv.org/abs/1908.10084}, 
}

@misc{kingma2017adammethodstochasticoptimization,
      title={Adam: A Method for Stochastic Optimization}, 
      author={Diederik P. Kingma and Jimmy Ba},
      year={2017},
      eprint={1412.6980},
      archivePrefix={arXiv},
      primaryClass={cs.LG},
      url={https://arxiv.org/abs/1412.6980}, 
}

@misc{UCBCLTC2025,
    author={Derek Manky and Gil Baram},
    title={Beyond Phishing: Exploring the Rise of AI-enabled Cybercrime},
    url={https://cltc.berkeley.edu/2025/01/16/beyond-phishing-exploring-the-rise-of-ai-enabled-cybercrime/},
    year={2025},
    month={January},
    note={Accessed: 12-07-2025}
}

@misc{anthropic2025cyberesp,
    author={Anthropic},
    title={Disrupting the first reported AI-orchestrated cyber espionage campaign},
    url={https://www.anthropic.com/news/disrupting-AI-espionage},
    year={2025},
    month={November},
    note={Accessed: 12-07-2025}
}

@misc{cert_ua_2025,
    author={CERT-UA},
    title={ UAC-0001 cyberattacks on the security and defense sector using the LAMEHUG software tool, which uses LLM (large language model) (CERT-UA\#16039)},
    url={https://cert.gov.ua/article/6284730},
    year={2025},
    month={July},
    note={Accessed: 12-07-2025}
}

@inproceedings{eftimie2021case,
  title={A case study in anticipating insider vulnerabilities using psychological profiling},
  author={Eftimie, Sergiu and Cotenescu, Vlad and Moinescu, Radu and R{\u{a}}cuciu, Ciprian and Gl{\u{a}}van, Dragoș},
  booktitle={2021 IEEE International Black Sea Conference on Communications and Networking (BlackSeaCom)},
  pages={1--4},
  year={2021},
  organization={IEEE}
}

@inbook{Ruohonen_2025_InsiderThreat,
   title={What Do We Know About the Psychology of Insider Threats?},
   ISBN={9783031893636},
   ISSN={1867-822X},
   url={http://dx.doi.org/10.1007/978-3-031-89363-6_11},
   DOI={10.1007/978-3-031-89363-6_11},
   booktitle={Digital Forensics and Cyber Crime},
   publisher={Springer Nature Switzerland},
   author={Ruohonen, Jukka and Saddiqa, Mubashrah},
   year={2025},
   pages={186–211} }

@inproceedings{
jones2022capturing,
title={Capturing Failures of Large Language Models via Human Cognitive Biases},
author={Erik Jones and Jacob Steinhardt},
booktitle={Advances in Neural Information Processing Systems},
editor={Alice H. Oh and Alekh Agarwal and Danielle Belgrave and Kyunghyun Cho},
year={2022},
url={https://openreview.net/forum?id=fcO9Cgn-X-R}
}

@misc{park2023generativeagentsinteractivesimulacra,
      title={Generative Agents: Interactive Simulacra of Human Behavior}, 
      author={Joon Sung Park and Joseph C. O'Brien and Carrie J. Cai and Meredith Ringel Morris and Percy Liang and Michael S. Bernstein},
      year={2023},
      eprint={2304.03442},
      archivePrefix={arXiv},
      primaryClass={cs.HC},
      url={https://arxiv.org/abs/2304.03442}, 
}
